\documentclass{bmvc2k}

\title{Mixup for Test-Time Training}

\addauthor{Bochao Zhang}{csbczhang@comp.hkbu.edu.hk}{1}
\addauthor{Rui Shao}{rui.shao@ntu.edu.sg}{2}
\addauthor{Jingda Du}{csjddu@comp.hkbu.edu.hk}{1}
\addauthor{PC Yuen}{pcyuen@comp.hkbu.edu.hk}{1}

\addinstitution{
 Hong Kong Baptist University\\
 Hong Kong SAR, China
}
\addinstitution{
 Nanyang Technological University\\
 Singapore
}

\runninghead{ }{ }

\begin{document}

\maketitle

\begin{abstract}
Test-time training provides a new approach solving the problem of domain shift. In its framework, a test-time training phase is inserted between training phase and test phase. During test-time training phase, usually parts of the model are updated with test sample(s). Then the updated model will be used in the test phase. However, utilizing test samples for test-time training has some limitations. Firstly, it will lead to overfitting to the test-time procedure thus hurt the performance on the main task. Besides, updating part of the model without changing other parts will induce a mismatch problem. Thus it is hard to perform better on the main task. To relieve above problems, we propose to use mixup in test-time training (\textbf{MixTTT}) which controls the change of model's parameters as well as completing the test-time procedure. We theoretically show its contribution in alleviating the mismatch problem of updated part and static part for the main task as a specific regularization effect for test-time training. \textbf{MixTTT} can be used as an add-on module in general test-time training based methods to further improve their performance. Experimental results show the effectiveness of our method.
\end{abstract}

\section{Introduction}
\label{sec:intro}
Usually in deep learning based methods, training phase and test phase are strictly separated. E.g. a normal dataset will be divided into three parts:training set, validation set and test set. Models are trained with training set data, while hyperparameters of the model are chosen with respect to the performance on validation set. Then the well trained model is evaluated on the test set. However, the demand of a good model is not restricted to good performance on the test set from the same dataset but also from other datasets. Namely, under distribution shifts whether the model still enjoys a high generalization ability and keeps its high performance. This problem is specifically studied in many fields with different settings like  domain adaptation, domain generalization, adversarial learning etc. Test-time training opens a new learning paradigm and provides a new strategy to counter this problem. 

Test-time training inserts a test-time training phase between the normal training phase and test phase. The model would be well trained in the training phase. Then when a test sample(s) comes for inference, it carries in itself information that could be utilized in test-time training phase. Such information includes domain information and visual information etc. Usually test-time training phase will fine-tune the model or important statistics (e.g. prototypes) as completing the auxiliary unsupervised task. Finally inference on the main task will be performed with the updated model (statistics) in test phase. From the setting view, test-time training obeys the rule of no access to test data during training. It coincides with domain generalization. From the view of test-time training process, whole or part of the model is updated to adapt to the test sample. In this way it coincides with domain adaptation. So test-time training basically puts very little requirement for the data and model, yet can counter certain degree of domain shift in its process.   
The early network structure of test-time training from \cite{ttt} is a multitask training framework. One main task and one self-supervised auxiliary task share a feature extractor and each task has an independent classifier, which is called head. Self-supervised task rotation~\cite{rotation} is chosen as auxiliary task for the test-time procedure. Simple multi-task training is used for the first phase. In test-time training phase of \cite{ttt}, a test sample(s) will be rotated to perform the auxiliary task and update the feature extractor. In test phase inference will be performed on the updated feature extractor and the original classifier. This network structure is considered as one of the most classical network structures for test-time training methods. Therefore, we conduct our theoretical analysis about MixTTT and ordinary TTT based on this framework. 

Test-time training aims that optimizing auxiliary task will intermediately improve the main task through the updated shared model. This could be realized with two approaches.
The first one relies that auxiliary task can minimize the domain discrepancy between the training set and test samples. Naturally when performing the main task, more accurate results could be obtained. Most test-time training methods follow this approach. However in such cases, a batch of test samples are demanded. 
The second approach depends on the cooperation of the main task and auxiliary task on certain datasets. It is expected that auxiliary task can help dig the inherent properties of the test sample. E.g. by performing the auxiliary task, visual information of test sample could be better extracted. Under such situation, optimizing auxiliary task will intermediately optimize the main task. This approach normally does not restrict the number of test samples. However it requires more delicate test-time update process to keep the good relation of two tasks, especially on unseen samples. As we mentioned above, uncontrolled optimization on the auxiliary task will cause overfitting. Besides, with much change of some parts of the model the static part for the main task will not work well on top of it. 

In this paper we propose to utilize mixup between training data and the test sample(s) to mitigate the above problems in test-time procedure. Our method can be applied on both the first approach setting and the second approach setting without specific requirement about the number of test sample(s). As a result, our add-on module allows test-time training to improve the main task with the full strength. In summary, our contribution is three-folded: \begin{itemize}
\item We identify an important problem in test-time training, which is model mismatch between the updated part and static part when accomplishing the auxiliary task. 
\item We show from theoretical analysis that the effect using mixup in test-time training brings implicit control in model change beyond original test-time training.
\item MixTTT can be seen as an add-on module without specific requirement about the number of test samples which can further boost the performance of existing test-time training related methods.   
\end{itemize}

\newpage

\section{Related Work}
\subsection{Test-time training}
Test-time training~\cite{ttt} opens a new learning paradigm to solve domain shift problem. In its belief, when the test sample comes for inference, it is a waste to not explore it. The core idea is to utilize the test sample information to optimize some parts of the model with the unsupervised auxiliary task and then evaluate the test sample with the updated model for the main task. Following its idea, appears test-time adaptation, for which usually many test samples are required to perform adaptation. Under the setting of test-time adaptation, many domain adaptation related techniques could be borrowed and utilized. Namely in such methods the aim of auxiliary task is to minimize the discrepancy between the source and target domains. \cite{wang2020tent} estimates normalization statistics and updates affine transformations to reduce entropy in the auxiliary task. \cite{ttt++} 
performs contrastive learning on the test sample batch with its augmented versions and on the other hand aims to align the mean and variance of features from the test sample batch with the training set. For now most test-time methods belong to test-time adaptation, there is rather limited paper solving single-test-sample based test-time training problem. Our method can fit in with both multiple-test-sample-based and single-test-sample-based test-time procedure.

Test-time training itself could also be used as an add-on module in domain generalization methods to improve the performance. 
\cite{ttclassifier} does not literally update the model during test, instead it chooses to update prototypes in a memory bank. At the end, the test sample is classified based on the distance of its feature vector with adjusted prototype representations.  
\cite{DGvia} does not explicitly mention test-time training, but they share similar idea in utilizing the test sample to do optimization for better inference. It aims to project the target sample to the source domain manifold through an inference-time procedure thus get more accurate inference outcome. 

Approaches from other fields like medical semantic segmentation \cite{segmentation1,segmentation2} and face anti-spoofing \cite{faceAS1,faceAS2} also utilize the idea of test-time training together with specific domain knowledge to solve the domain shift problem.

\subsection{Mixup}
The core idea of mixup~\cite{mixup} is that convex combination of sample pair and their labels forms a general vicinal distribution. Previous experience shows that samples drawn from vicinal distribution increase the amount of training samples and relieve overfitting. 

\cite{manifoldmixup} claims that manifold mixup as a regularization method gives smoother decision boundary and better regularization during training stage.    
More mixup-based augmentation methods gradually appear as well e.g. \cite{cutmix} further improves the performance on localization. \cite{puzzlemix} gives a more effective strategy on how to cut and mix sample pairs. 

Mixup as a data processing method is also used to solve domain generalization and domain adaptation problems. \cite{DGdomainmixup} separates two kinds of mixup: mixing samples from two different domains and mixing samples from all domains. The second one shows good performance in visual decathlon benchmark~\cite{vd}. \cite{DAmixup1,DAmixup2} both utilize the concepts of mixup and adversarial training for domain adaptation but in different ways. \cite{DAmixup1} chooses to mix source domain data with the target domain data, thus filling the gap with mixed samples between two domains. \cite{DAmixup2} instead mix up samples within the source domain and the target domain. 

There also exist papers proving efficacy of mixup through theoretical analysis. e.g. \cite{howmixup} expands the mixup loss through Taylor expansion and boils it down to the standard loss plus some regularization terms. It explicitly shows mixup improves robustness with respect to certain adversarial attacks and exhibits data-adaptive regularization effect for generalization. Our method, however, explains the regularization effect of mixup during test-time training from the perspective of stabilizing the feature extractor thus maintaining the good cooperation of it with classifier.  

\section{Test-Time Training under Model Mismatch}
In this section, we will first describe the rise of model mismatch and its influence in test-time training then give a strategy solving this problem. We explore the mixup based test-time training loss (MixTTT loss)  and ordinary test-time training loss. Through Taylor expansion it can be noticed that besides ordinary loss, MixTTT loss includes a regularization term that constrains the model update for test-time procedure.

\subsection{Model Mismatch}
Test-time training utilizes information carried in the test sample before its final inference. During this procedure, parts of the model are updated while other parts remain static (untouched). This procedure is dynamic and there exist samples showing the following property. As completing the auxiliary task, the information contained in the test sample is initially utilized in the positive direction. With the different designs of auxiliary task, the positive effect could be letting the model be more sensitive to the the visual structure of the test sample or minimizing the domain discrepancy between the training data and test data etc. However, 
when the test-time procedure goes further, the model would be overfitting to the auxiliary unsupervised task thus neglecting the true goal of test-time training: the main task. One obvious overfitting consequence is model mismatch between the updated part and static part for the main task. 
\begin{figure}[h]
\centering
\includegraphics[width=\linewidth]{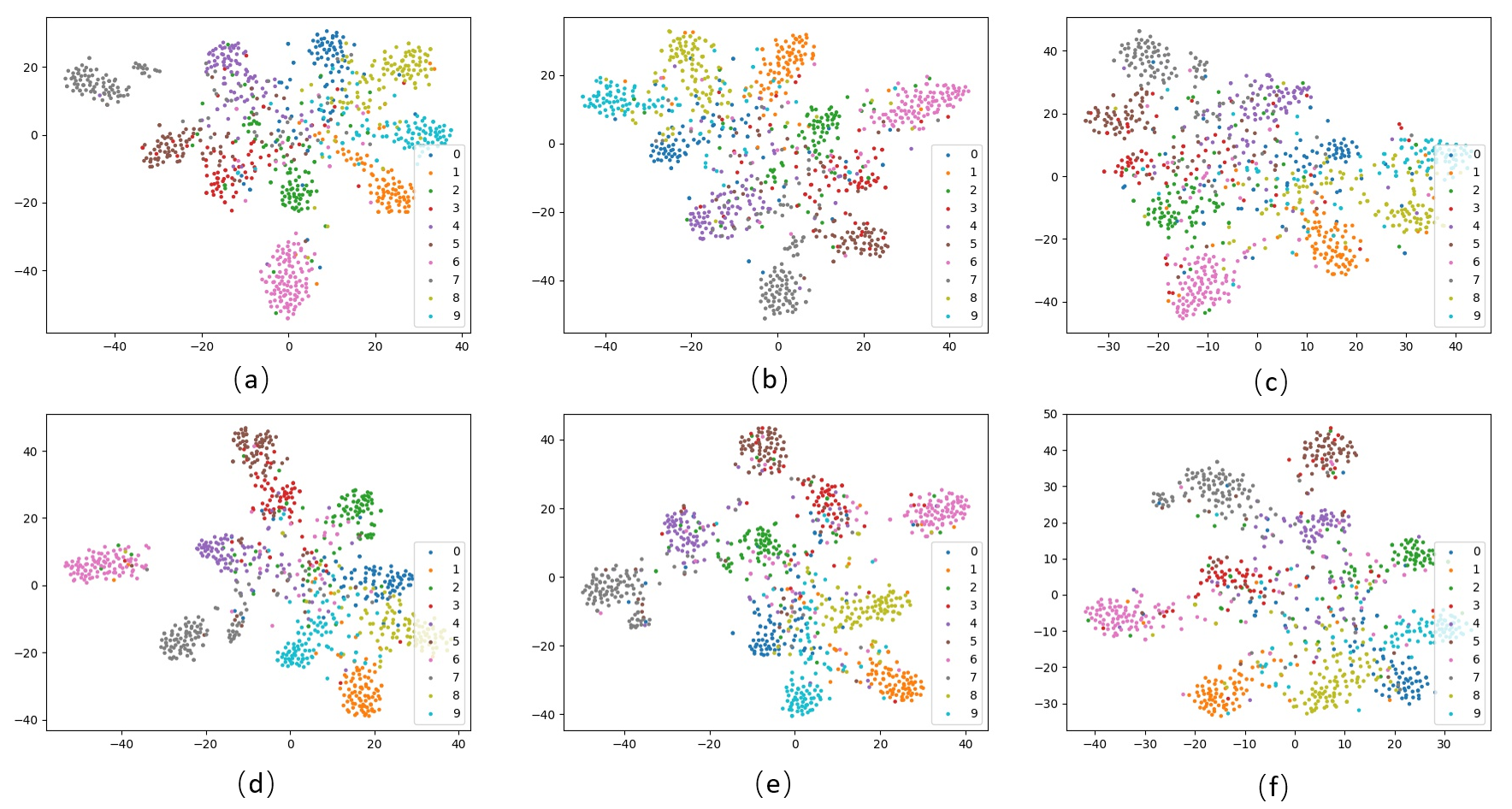}
\caption{t-SNE visualization of embedding space of 1000 sample points}
\label{vis}
\end{figure}

In order to witness this mismatch problem more clearly, we visualize the embedding space of 1000 test sample points from CIFAR-10-C\cite{corruption} with the \cite{ttt} framework and see its change during different test-time training (TTT) stage. Overall we want to show that it's hard to use the original main task head (classifier) to classify features from an overly changed feature extractor and MixTTT makes this task easier. Figure 1(a), 1(b), 1(c) show the embedding space variation of 1000 test samples after performing ordinary TTT. Each of them shows the embedding space after 10, 20, 30 steps of iteration in the test-time procedure. Figure 1(d), 1(e), 1(f) show the embedding space of the same data points after performing MixTTT with each corresponding to 10, 20, 30 steps of iterations during the test-time optimization. All above whole test-time procedure utilizes the same learning rate 1e-3.  
It can be noticed that in the embedding space of ordinary TTT clusters are initially separated as shown in Figure 1(a) with each class basically forming one cluster. As the test-time procedure goes, the clusters become chaotic and start to diffuse into other clusters as shown in Figure 1(b). In Figure 1(c) the embedding space is more messy, which puts more difficulty for the original classifier to distinguish these samples. MixTTT rather produces a less chaotic embedding space, which will ease the burden for classifier. In a nutshell, with an overly changed feature extractor, clusters are scattered and overlapped. Thus original classifier is not suitable to classify such messy embedding space. This problem is considered as model mismatch in test-time training.    

\subsection{Mixup for Test-Time Training}
For this section, we will briefly give the mixup based test-time training approach (\textbf{MixTTT}) and show how it can be used as an add-on module on existing test-time training based methods. Next we prove the efficacy of MixTTT in controlling the model change during test-time training with the framework from \cite{ttt}. 

For each test-time optimization step, test sample(s) will be mixed up with randomly selected training samples using different mixup ratio drawn from Uniform distribution and form a batch of mixed data. Here fixed training set samples and fixed mixup ratio will all deteriorate the diversity of mixed samples and reduce fusion of the test sample with the training set distribution and thus restrict the regularization effect of mixup based test-time training. Then the mixed data is to perform auxiliary task optimization according to the setting of each test-time training method. We give an example of TTT-R\cite{ttt} with MixTTT. The whole process is shown in Algorithm \ref{alg:mixttt}.

Let us define the notations of different variables for convenience. We consider a training set $S_{tr}=\{(x_{1},y_{1}),(x_{2},y_{2}),...(x_{i},y_{i})...,(x_{n},y_{n})\}$, where $x_{i}$ represents one training sample and $y_{i}$ the corresponding main task label. 
One single test is denoted as $x_{t}$ and its corresponding auxiliary self-supervised task label $y_t$. We denote $\lambda$ as the mixup ratio and one mixed sample $\tilde{x}_{i,t}=\lambda x_{i}+(1-\lambda ) x_{t}$. Let $f$ denote the shared feature extractor and $hm$, $hs$ two independent classifiers of the main task and auxiliary self-supervised task.
We package the whole auxiliary model as one function, denoted as $g_{\theta }$. We denote $L_{mt}$ as the loss with mixed samples as input in test-time training and $L_{t}$ as the original test-time training loss. $C$ is the function calculating auxiliary self-supervised loss.

\begin{algorithm}[h]
\caption{MixTTT}
\label{alg:mixttt}
\textbf{Input}: one test-time training based method f; single test sample $x_{t}$, training set images,$\{x_{1},x_{2}...x_{n}\}$, shared feature extractor $f_{\theta}$, head for main task $hm_{\phi1}$, head for auxiliary task $hs_{\phi2}$ , Loss function for auxiliary task $C$\\
\textbf{Parameter}: learning rate $\alpha$, mixup ratio list $\lambda_L$\\
\textbf{Output}: inference result
\begin{algorithmic}[1] 
\WHILE{$iteration step \leq  total steps$}
\STATE sample a mixup ratio $\lambda$
\STATE sample a batch of training images $X$
\FOR{each training sample in the batch}
\STATE mix the training sample with the test sample and get the mixed batch $X_m$
\ENDFOR 
\STATE calculate auxiliary loss with the following equation: 
    \begin{equation}
    L_{aux}=C(X_m;hs_{\phi2},f_{\theta})
    \end{equation}
    \vspace*{-0.5cm}
\STATE update the shared feature extractor as follows:
    \begin{align}
        f_{\theta} \leftarrow f_{\theta} - \nabla_{\theta } L_{aux}
    \end{align}
    \vspace*{-0.5cm}
\ENDWHILE
\STATE inference on the test sample with the following equation:
    \begin{align}
        \hat{y}_t = hm_{\phi1} \circ f_{\theta}(x_t)
    \end{align}
    \vspace*{-0.5cm}
\STATE \textbf{return} inference result
\end{algorithmic}
\end{algorithm}

From the start of equation (4) we show that how MixTTT loss $L_{mt}$ is beyond original test-time training loss $L_{t}$. The two loss functions are shown in equation (4) and (5). Here we consider the classification as the auxiliary self-supervised task and its corresponding label $y_t$.
\begin{equation}
L_{t}=-y_{t}^\intercal \cdot log(g_{\theta }(x_{t})) 
\end{equation}
\begin{equation}
    L_{mt}=-y_{t}^\intercal \cdot log(g_{\theta }(\tilde{x}_{i,t}))=-y_{t}^\intercal \cdot log(g_{\theta }(\lambda\cdot x_{i}+(1-\lambda )\cdot x_{t}))
\end{equation}

Next we write the mixup loss with Taylor expansion at mixup ratio $\lambda$ equals 0. 
\begin{equation}
\begin{split}
        L_{mt} = -y_{t}^\intercal \cdot log(g_{\theta }(x_{t}))+ \nabla_{\lambda} L_{mt}(0)\lambda+ \mathcal{O}(\lambda^{2})    =L_{t}+ \nabla_{\lambda} L_{mt}(0)\lambda+ \mathcal{O}(\lambda^{2})
\end{split}
\end{equation}

From equation (6) we notice that mixup in test-time training in its loss function provides more regularization terms besides original test-time training loss. Now we focus on the regularization term and denote it as $L_r$.
\begin{equation}
    L_r=\nabla_{\lambda} L_{mt}(0)\lambda
\end{equation}

\begin{equation}
    \nabla_{\lambda} L_{mt}(\lambda) = -\{(y_t \odot g_\theta (\tilde{x}_{i,t})_{-1})^\intercal [ \frac{\partial g_\theta (x)}{\partial x}^\intercal (x_i-x_t)] \}^\intercal
\end{equation}
From another perspective, we can write the gradient of original test-time training loss over input x as follows:
\begin{equation}
\setlength{\abovedisplayskip}{1pt}
\setlength{\belowdisplayskip}{1pt}
\nabla_{x} L_{t}(x) = -\{(y_t \odot g_\theta (\tilde{x}_{i,t})_{-1})^\intercal  \frac{\partial g_\theta (x)}{\partial x}^\intercal  \}^\intercal
\end{equation}
Since we expand the loss at $\lambda$ equals zero, then $x$ equals $x_t$. We can get $L_r=(x_i-x_t)^\intercal \nabla_{x} L_{t}(x_t) \lambda$ and $L_{mt}$ as follows:
\begin{equation}
    L_{mt} = L_{t}+ (x_i-x_t)^\intercal \nabla_{x} L_{t}(x_t) \lambda +\mathcal{O}(\lambda ^{2})
\end{equation}
\begin{equation}
    \nabla_x{L_t}=\frac{\partial feat}{\partial x}\frac{\partial L_t}{\partial feat}  
\end{equation}
\begin{equation} 
    \|\nabla_\theta{L_t}\|= \|\frac{\partial feat}{\partial \theta}\frac{\partial L_t}{\partial feat}\|
\end{equation}

 Equation (11) is to measure how much the feature extractor is supposed to change in back propagating such loss. By manually select $x_i$ bigger than $x_t$, we minimize the loss is to minimize $\nabla_x{L_t}$. Since feature is determined by two factors $\theta$ and $x$, for back propagation $x$ is fixed and will not be updated. So the derivative of feature over x is fixed. So minimizing $\nabla_x{L_t}$ minimizes the $\frac{\partial L_t}{\partial feat}$ which is a shared part of $\|\nabla_\theta{L_t}\|$. Thus the change of feature extractor is intermediately minimized in Equation (11).

\section{Experiments}
\subsection{Datasets and Experimental Settings}
We evaluate the effectiveness of MixTTT on existing test-time training based methods following their own test-time auxiliary task. Our chosen methods are TTT-R~\cite{ttt}, Tent~\cite{wang2020tent} and TTT++\cite{ttt++}. We also choose CIFAR-10-C and CIFAR-100-C~\cite{corruption} for evaluation. The dataset setting of CIFAR-10-C and CIFAR-100-C are similar in many ways. They both contain 15$\times$5$\times$10000 colour images and the size of all images is 32$\times$32. 15 corresponds to 15 different kinds of corruption and 5 corresponds to 5 levels of severities. These 15 kinds of corruption reflects common situations that images will be under e.g. weather change, different lighting conditions, compression etc. Therefore it can widely verify the robustness of algorithms. We choose the most severe level to carry experiments. CIFAR-100-C is a 100-way classification dataset, while CIFAR-10-C is a 10-way classification dataset. Since the difference lies in the difficulty for these two datasets, their experimental results actually verify the problem we identified, which will be discussed in the next part. 

We introduce the experimental setting of test-time procedure with MixTTT. In TTT-R~\cite{ttt}, we train a network with ResNet50 as the backbone. Single test sample is to mixup with training samples and form a batch of mixed samples. These mixed samples are then rotated with a certain degree. and then to perform the auxiliary task. Considering the iteration step is limited, mixup ratio is selected with bigger proportion for the test sample, so we sample it from a uniform distribution $U[0.7, 1]$. The learning rate for CIFAR-10-C is set as 1e-3 and the iteration steps is 10. For CIFAR-100-C learning rate is set as 1e-4 and the iteration steps is 5. In Tent~\cite{wang2020tent} test samples are to mix with training samples. Since the label for test samples are unknown, we choose to mix them with high proportion test images to not severely disturb the class label. We sample the mixup ratio from $U[0.95, 1]$. The mixed images are to perform entropy minimization to adjust the affine parameters in batch normalization layers. The experimental results demonstrate that MixTTT influence the test-time procedure in a positive way. We implement it using the tent code part from \cite{ttt++} and borrow the well trained checkpoints and hyperparameter setting from it for the test-time procedure. In TTT++\cite{ttt++} the test-time procedure is to perform contrastive learning and feature alignment. When applying MixTTT in contrastive learning, the goal is set as maximizing the agreement between augmented test images and mixed images. To maintain the class label of test images, the mixup ratio is set for bigger proportion test images $U[0.9, 1]$. We also use the setting, code and checkpoints from \cite{ttt++}. All the experiments are carried on a Nvidia V100 GPU. 

\subsection{Experimental Results}
Table \ref{tab:cifar10} shows the error rate of existing methods and its application with MixTTT on CIFAR-10-C. The initial model checkpoints for corresponding existing methods and its application with MixTTT are the same for a fair comparison. It can be noticed that MixTTT helps improve the main task performance with different auxiliary tasks in test-time procedure e.g. rotation, entropy minimization and contrastive learning. Although mixup may introduce some bias for entropy minimization, within a range it could still counter model mismatch and benefit the test-time procedure. For CIFAR-10-C, test-time procedure with MixTTT performs better than original test-time procedure for almost all kinds of corruption. For CIFAR-100-C, test-time procedure with MixTTT performs better than original test-time procedure for all kinds of corruption. 
It can be also noticed from Table \ref{tab:cifar100} that for different kinds of unsupervised test-time procedure, our add-on module is effective and helpful and can suit situations with different number of test samples. 

Overall the 10-way classification on CIFAR-10-C is easier than the task on CIFAR-100-C which has 100 classes. The difference is not just the number of classes. If using the same backbone for feature extraction, then the load of classifier for 100-way classification is heavier than 10-way classification. That is to say, 100-way classification is prune to be influenced by model mismatch than 10-way classification. This also explains task on CIFAR-100-C benefits more than task on CIFAR-10-C.
\begin{table*}[h]
\centering
\resizebox{\linewidth}{!}{
\begin{tabular}{lrrrrrrrrrrrrrrrrr}
\toprule
Methods & gauss & shot &  impul  & defoc  &  glass & motn & zoom &  snow  & frost & fog & brit  & contr  & elast  &  pixel  & jpeg  & avg  \\ 
\midrule
TTT-R\cite{ttt} & 26.42 & 23.77 & 33.50 & 12.66 & 31.02 & 17.36 & 12.36 & 15.06 & 15.60 & 19.75 & 9.40 & 16.08 & 23.79 & 21.50 & 22.83 & 20.07  \\
SHOT\cite{shot} & 15.66 & 14.67 & 24.84 & 8.96 & 23.13 & 12.42 & 7.42 & 14.07 & 12.42 & 17.06 & 7.79 & 7.67 & 17.88 & 10.68 & 12.98 & 13.84\\
Tent\cite{wang2020tent} & 14.47 & 13.10 & 21.80 & 8.34 & 21.08 & 11.06 & 6.98 & 12.06 & 11.71 & 14.04 & 7.11 & 7.02 & 16.70 & 10.19 & 12.05 & 12.51  \\ 
TTT++\cite{ttt++} & 14.10 & 12.28 & 12.12 & 8.74 & \textbf{16.98} & 9.91 & 7.17 & 10.15 & 10.60 & 9.22 & 5.84 & 6.24 & 14.04 & 16.98 & 11.59 & 11.06  \\ \hline
\cite{ttt}+MixTTT & 25.32 & 22.54 & 32.67 & 11.60 & 29.51 & 16.00 & 11.30 & 15.01 & 14.34 & 18.16 & 8.77 & 8.99 & 22.34 & 20.82 & 21.81 & 18.61\\ 
\cite{wang2020tent}+MixTTT & 14.28 & 12.99 & 22.09 & 8.27 & 20.57 & 10.83 & 7.14 & 11.52 & 11.52 & 13.05 & 6.98 & 7.84 & 16.27 & 9.69 & 12.05 & 12.33  \\ 
\cite{ttt++}+MixTTT & \textbf{12.72} & \textbf{11.55} & \textbf{11.85} & \textbf{7.96} & 17.04 & \textbf{8.95} & \textbf{6.97} & \textbf{9.12} & \textbf{9.37} & \textbf{8.45} & \textbf{5.36} & \textbf{5.43} & \textbf{13.50} & \textbf{8.88} & \textbf{11.06} & \textbf{9.88}\\ 
\bottomrule
\end{tabular}}
\vspace*{0.0cm}
\caption{Error rate (\%) on CIFAR-10-C with level 5 corruption.}
\label{tab:cifar10}
\end{table*}
\begin{table*}[h]
\centering
\resizebox{\linewidth}{!}{
\begin{tabular}{lrrrrrrrrrrrrrrrrr}
\toprule
Methods & gauss & shot &  impul  & defoc  &  glass & motn & zoom &  snow  & frost & fog & brit  & contr  & elast  &  pixel  & jpeg  & avg  \\ 
\midrule
TTT-R\cite{ttt} & 78.07 & 75.79 & 89.34 & 32.99 & 79.37 & 46.22 & 32.45 & 48.21 & 52.06 & 54.52 & 26.50 & 45.16 & 53.45 & 65.78 & 53.83 & 55.58  \\
SHOT\cite{shot} & 42.73 & 40.89 & 52.20 & 31.04 & 50.23 & 35.99 & 29.18 & 42.08 & 39.27 & 45.60 & 29.38 & 29.86 & 43.19 & 33.45 & 35.75 & 38.72 \\
Tent\cite{wang2020tent} & 40.45 & 38.55 & 48.50 & 29.03 & 46.78 & 32.71 & 28.02 & 38.50 & 36.64 & 37.86 & 27.60 & 27.93 & 40.44 & 31.47 & 34.33 & 35.92  \\ 
TTT++\cite{ttt++} & 36.76 & 34.76 & 39.90 & 27.98 & 40.67 & 29.75 & 25.86 & 31.71 & 33.42 & 32.24 & 24.11 & 25.03 & 35.72 & 29.58 & 31.82 & 31.95  \\ \hline
\cite{ttt}+MixTTT & 72.45 & 68.91 & 80.23 & 35.93 & 72.73 & 45.17 & 34.32 & 46.34 & 47.93 & 52.87 & 28.80 & 44.90 & 53.24 & 61.39 & 52.31 & 53.16  \\ 
\cite{wang2020tent}+MixTTT & 40.17 & 38.64 & 48.50 & 28.85 & 46.44 & 32.18 & 27.42 & 37.99 & 36.55 & 36.96 & 27.24 & 29.56 & 40.47 & 31.09 & 33.79 & 35.72  \\ 
\cite{ttt++}+MixTTT & \textbf{34.72} & \textbf{32.58} & \textbf{39.14} & \textbf{26.84} & \textbf{39.29} & \textbf{28.54} & \textbf{24.94} & \textbf{31.42} & \textbf{31.83} & \textbf{30.80} & \textbf{23.54} & \textbf{23.58} & \textbf{30.09} & \textbf{27.92} & \textbf{30.30} & \textbf{30.36}  \\ 
\bottomrule

\end{tabular}}
\vspace*{.0cm}
\caption{Error rate (\%) on CIFAR-100-C with level 5 corruption.}

\label{tab:cifar100}
\end{table*}
\section{Conclusion and Discussion}
Through analyzing the whole test-time training process, we witness model mismatch as one of the biggest problems TTT is facing. We propose MixTTT as an add-on module for test-time training based methods and show from theoretical view how mixup in test-time training controls the model change thus further control model mismatch while completing the test-time auxiliary task. This add-on module can suit both the single-test-sample-based and multiple-test-sample-based test-time procedure following the ordinary optimization process in such procedure.

Model mismatch is not the only problem for test-time training. As the two approaches we mentioned above, how to dig and utilize the information from test sample(s) is worth more attention and could deliver extra gain for the main task. The development of more informative unsupervised or self-supervised task will further improve test-time training. The development of training mechanisms that can entangle the auxiliary task and the main task more closely will also improve test-time training related methods. 

Besides, since test-time training is a new paradigm, unified evaluation concerning computation cost, test-time training time, access of data should be proposed and followed.  

\bibliography{egbib}

\end{document}